\documentclass[sigconf]{acmart}
\renewcommand\footnotetextcopyrightpermission[1]{} 
\setcopyright{none}
\AtBeginDocument{%
  }

\setcopyright{acmlicensed}
\copyrightyear{2026}
\acmYear{2026}
\acmDOI{XXXXXXX.XXXXXXX}
\acmConference[Conference acronym '34]{Make sure to enter the correct
  conference title from your rights confirmation email}{November 10--14,
  2026}{Rio de Janeiro, Brazil}
\acmISBN{978-1-4503-XXXX-X/2026/06}
\usepackage{multirow} 
\usepackage{fontawesome}
\usepackage{xcolor}
\usepackage[linesnumbered,ruled,vlined]{algorithm2e}
\acmSubmissionID{4906}

\newcommand{\sysname}{\textbf{BARD}}
\newcommand{\modname}{\textbf{BARD-VL}}
\definecolor{DarkGreen}{RGB}{0,100,0}
\definecolor{lightblue}{RGB}{95, 158, 160} 
\begin{document}


\title{BARD: Bridging AutoRegressive and Diffusion Vision-Language \\ Models Via Highly Efficient Progressive Block Merging and Stage-Wise Distillation}

\author{Baoyou Chen$^{1,3,\star}$ \quad HanChen Xia$^{1,\star}$ \quad Peng Tu$^{1,\star}$ \quad
        Haojun Shi$^{1}$ \quad Liwei Zhang$^{1}$  \quad Yuxuan Yao$^{2,3}$ \quad Weihao Yuan$^{4}$ \quad Siyu Zhu$^{1,2,3,\dagger}$}
\authornote{$^\star$ Equal contribution. $^\dagger$ Corresponding author.}
\affiliation{%
  \institution{$^{1}$Shanghai Academy of AI for Science, $^{2}$Shanghai Innovation Institute, $^{3}$Fudan University}
  \institution{$^{4}$Nanjing University}
  \country{}
}

\renewcommand{\shortauthors}{}

\begin{abstract}
Autoregressive vision-language models (VLMs) deliver strong multimodal capability, 
but their token-by-token decoding imposes a fundamental inference bottleneck.
Diffusion VLMs offer a more parallel decoding paradigm, 
yet directly converting a pretrained autoregressive VLM into a large-block diffusion VLM (dVLM) often leads to substantial quality degradation.
In this work, we present \sysname, 
a simple and effective bridging framework that converts a pretrained autoregressive VLM into a same-architecture, decoding-efficient dVLM.
Our approach combines progressive supervised block merging, 
which gradually enlarges the decoding block size, with stage-wise intra-dVLM distillation from a fixed small-block diffusion anchor to recover performance lost at larger blocks.
We further incorporate a mixed noise scheduler to improve robustness and token revision during denoising, 
and memory-friendly training to enable efficient training on long multimodal sequences.
A key empirical finding is that direct autoregressive-to-diffusion distillation is poorly aligned and can even hurt performance, whereas distillation within the diffusion regime is consistently effective.
Experimental results show that, with $\leq 4.4M$ data, \modname\ transfers strong multimodal capability from Qwen3-VL to a large-block dVLM. Remarkably, \modname\ establishes a new SOTA among comparable-scale open dVLMs on our evaluation suite at both 4B and 8B scales. At the same time, \modname\ achieves up to \textbf{3$\times$} decoding throughput speedup compared to the source model. \textit{Code is available at: \textcolor{lightblue}{ https://github.com/fudan-generative-vision/Bard-VL}}.
\end{abstract}





\begin{teaserfigure}
    \centering
    \includegraphics[width=0.88\linewidth]{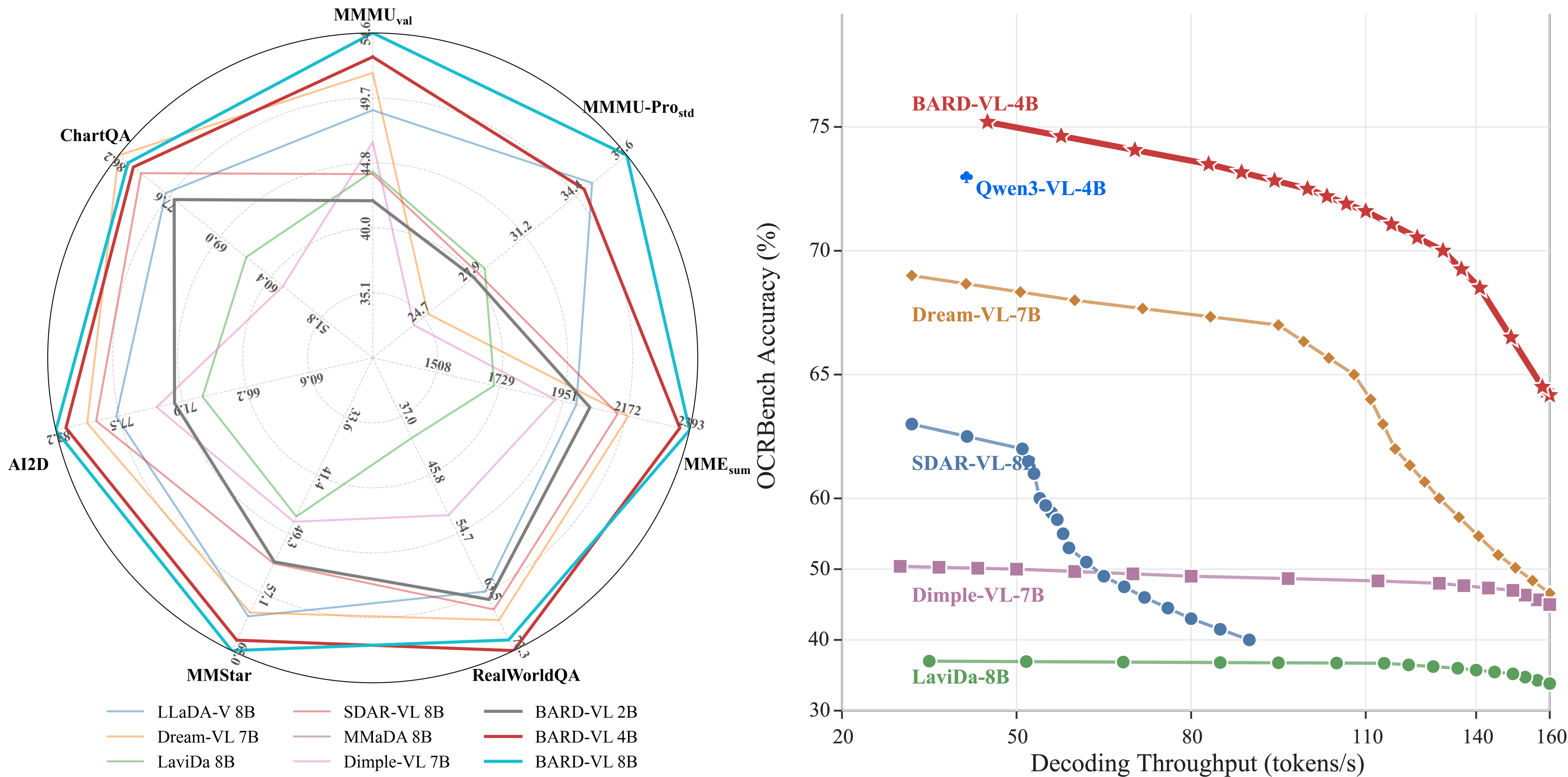}
    \caption{\textbf{Quality--efficiency comparison of \modname\ and representative open dVLMs.} Left: radar chart on seven multimodal benchmarks, where \modname\ at 2B/4B/8B is compared with prior open diffusion VLMs. \modname\ 4B and 8B exhibit the strongest overall performance among the compared dVLMs, showing that the proposed bridge substantially narrows the capability gap of existing diffusion VLMs. Right: OCRBench accuracy versus decoding throughput. Even though it is smaller than the 7B/8B baselines, \modname\ 4B traces a clearly better accuracy--throughput trade-off, retaining higher accuracy across a broad range of decoding speeds.}
    \label{fig:performance-radar-map}
\end{teaserfigure}

\maketitle

\section{Introduction}

Autoregressive vision-language models (VLMs) have become the dominant foundation for multimodal understanding and agentic interaction. Their success spans visual reasoning, document understanding, grounded question answering, and emerging multimodal agents. Yet their token-by-token causal decoding remains inherently sequential, imposing a hard inference bottleneck for practical deployment. Diffusion-based decoding offers a compelling alternative: by refining a partially generated response block by block, a diffusion VLM (dVLM) can update multiple tokens in parallel and therefore expose a fundamentally different quality--efficiency trade-off \cite{arriola2025block,zeng2025diffusionvl,li2025lavida}.

Recent progress has established diffusion models as a viable direction for both language and vision-language modeling. Broadly speaking, existing methods follow two main paradigms. The first builds diffusion-native models from scratch or from dedicated diffusion backbones, represented by MDLM and LLaDA on the language side, and Dimple, LaViDa, and LLaDA-V on the multimodal side \cite{sahoo2024simple,nie2025large,yu2505dimple,li2025lavida,you2025llada}. The second starts from strong autoregressive checkpoints and bridges them into diffusion models through post-training or continuous pretraining, as exemplified by Dream, NBDiff, Fast-dLLM v2, Efficient-DLM, and DiffusionVL \cite{ye2025dream,tian2025next,wu2025fast,fu2025efficient,zeng2025diffusionvl}. 
Recent work~\cite{xia2026t} studies the text-only bridging setting and does not address teacher-forcing post-training from a frontier autoregressive VLM to a strong large-block dVLM. 
Despite this rapid progress, current dVLMs cannot match comparable scale VLMs in overall multimodal capability.



In this work, we propose \sysname, 
an efficient framework for bridging a strong autoregressive VLM to a same-architecture large-block diffusion VLM. 
Our approach begins with the pretrained autoregressive checkpoint and first converts it into a stable small-block diffusion anchor. 
We then progressively increase decoding parallelism through progressive block merging, 
which smoothly expands the block size instead of forcing an abrupt transition from causal decoding to large-block diffusion. 
To recover the capability drop that arises at higher block sizes, 
we further introduce stage-wise dVLM distillation, 
where each larger-block model is distilled from the fixed small-block diffusion anchor.

Beyond the bridging procedure itself, 
we incorporate two practical improvements that make diffusion post-training more effective for multimodal models: 
a mixed noise scheduler, 
which trains the model to refine both masked and visibly corrupted tokens and thus improves iterative revision, 
and memory-friendly training, 
which packs clean and noisy responses with a shared multimodal context to significantly reduce training overhead. 
Together, these components form a simple and scalable recipe for teacher-forcing post-training of large-block dVLMs. 
Using fewer than 4.4M training samples, 
\sysname\ transfers the capability of a frontier autoregressive VLM to a highly parallel diffusion VLM with essentially no performance loss.

Using the proposed framework, we train a family of dVLMs, \modname, at multiple model scales. Experiments on multimodal reasoning, visual question answering, and document understanding benchmarks show that models trained with \modname\ effectively eliminate the performance gap between diffusion and autoregressive VLMs. At both 4B and 8B scales, \modname\ consistently outperforms prior open dVLM baselines and achieves the strongest results among comparable-scale dVLMs in our evaluation suite, while retaining, matching, or slightly exceeding the performance of its Qwen3-VL source model on most reported benchmarks. At the same time, \modname\ delivers up to \textbf{3$\times$} decoding throughput speedup compared to the source model. Figure~\ref{fig:performance-radar-map} summarizes the resulting quality--efficiency profile.
\section{Related Work}

\paragraph{\textbf{Discrete Diffusion Language Models.}}
Recent work on discrete diffusion models replaces causal next-token prediction with iterative denoising over discrete tokens. In language modeling, MDLM, LLaDA, and Dream show that simple mask-based training can already scale to strong performance \cite{sahoo2024simple,nie2025large,ye2025dream}. Subsequent studies mainly examine two issues that are also central to our setting: the choice of corruption process and the mismatch between random corruption during training and progressive generation at inference. GIDD broadens the noising process beyond pure masking \cite{von2025generalized}, while MDPO studies how training can be better matched to the generation process used at inference time \cite{he2025mdpo}.

\paragraph{\textbf{Discrete Vision-Language Models.}}
This framework has also been extended to multimodal settings. MMaDA studies multimodal diffusion foundation models, while Dimple, LaViDa, LLaDA-V, and DiffusionVL develop diffusion-based vision-language models for multimodal understanding \cite{yang2025mmada,yu2505dimple,li2025lavida,you2025llada,zeng2025diffusionvl}. RIV and ReDiff place more emphasis on iterative correction, and FUDOKI explores a flow-based alternative \cite{li2025riv,ji2025denoising,wang2025fudoki}. Across these works, a recurring difficulty is revision: once a wrong token appears early, later denoising steps often fail to correct it. Our mixed-noise training is designed to improve this part of the generation process.

\paragraph{\textbf{Autoregressive-to-Diffusion Adaptation.}}
A separate line of work asks whether strong pretrained autoregressive models can be converted into diffusion models through continued pretraining or post-training, rather than training diffusion models from scratch. In language modeling, several recent studies show that autoregressive pretraining provides a strong starting point for diffusion adaptation \cite{gong2024scaling,ye2025dream,wu2025fast,fu2025efficient}. The same idea has recently been extended to multimodal models. DiffusionVL presents a general recipe for converting autoregressive VLMs into diffusion VLMs, while SDAR-VL and Dream-VL show that this route can already reach strong performance on vision-language understanding benchmarks \cite{zeng2025diffusionvl,cheng2025sdar,ye2025dreamvl}.

Once the model is converted, the central question becomes how to increase parallelism without losing too much quality. Block Diffusion, Next-Block, and T$^\star$ show that quality degrades as the decoding block grows, and therefore adopt progressive block growth to make the transition to more parallel decoding more stable \cite{arriola2025block,tian2025next,xia2026t}. APD, ReFusion, and E2D2 address the same efficiency–quality trade-off from different angles, including inference scheduling, hybrid decoding, and model design \cite{israel2025accelerating,arriola2025encoder,luo2023refusion}. Our work is closest to the progressive block-scaling line, but studies it in autoregressive-to-diffusion VLM adaptation, where preserving multimodal capability is the main challenge.


\section{Methodology}
\label{sec:method}

\begin{algorithm}[t]
    \scriptsize
    \SetAlgoLined
    \SetAlCapSkip{0.5em}
    \setlength\belowcaptionskip{-5pt}
    \caption{\sysname: Progressive Supervised Block Merging with Stage-wise dVLM Distillation}
    \label{alg:pbm}

    \KwIn{Pretrained autoregressive VLM checkpoint \(\phi\), training set \(\mathcal D\), block schedule \(\{B_k\}_{k=1}^{K}\) with \((B_1,B_2,B_3,B_4)=(4,8,16,32)\), stage-training steps \(\{T_k\}_{k=1}^{K}\), stage-distillation steps \(\{S_k\}_{k=2}^{K}\)}
    \KwOut{Final dVLM checkpoint \(\theta^{(K)}\)}

    \tcp{Step 1: obtain the small-block anchor dVLM}
    \(\theta^{(1)} \leftarrow \phi\)\;
    \For{\(u \leftarrow 1\) \KwTo \(T_1\)}{
        sample minibatch \((Q,x_1)\) from \(\mathcal D\)\;
        sample corruption level \(t \sim \mathrm{U}(0,1)\) and \(x_t \sim q(x_t \mid x_1,t)\)\;
        update \(\theta^{(1)}\) using \(\mathcal L_{\mathrm{mix}}(\theta^{(1)}; B_1)\)\;
    }

    \tcp{Step 2: progressively enlarge the block size}
    \For{\(k \leftarrow 2\) \KwTo \(K\)}{
        \tcp{Stage training under block size \(B_k\)}
        \(\tilde{\theta}^{(k)} \leftarrow \theta^{(k-1)}\)\;
        \For{\(u \leftarrow 1\) \KwTo \(T_k\)}{
            sample minibatch \((Q,x_1)\) from \(\mathcal D\)\;
            sample corruption level \(t \sim \mathrm{U}(0,1)\) and \(x_t \sim q(x_t \mid x_1,t)\)\;
            update \(\tilde{\theta}^{(k)}\) using \(\mathcal L_{\mathrm{mix}}(\tilde{\theta}^{(k)}; B_k)\)\;
        }

        \tcp{Step 3: distill from the fixed anchor teacher \(\theta^{(1)}\)}
        \(\theta^{(k)} \leftarrow \tilde{\theta}^{(k)}\)\;
        freeze teacher \(\theta^{(1)}\)\;
        \For{\(s \leftarrow 1\) \KwTo \(S_k\)}{
            sample minibatch \((Q,x_1)\) from \(\mathcal D\)\;
            sample corruption level \(t \sim \mathrm{U}(0,1)\) and the same \(x_t \sim q(x_t \mid x_1,t)\) for teacher and student\;
            compute teacher logits \(z_T\) with \(\theta^{(1)}\)\;
            compute student logits \(z_S\) with \(\theta^{(k)}\)\;
            update \(\theta^{(k)}\) using \(\mathcal L_{\mathrm{kd}}^{(k)}(\theta^{(k)},\theta^{(1)})\)\;
        }
    }
    \KwRet{\(\theta^{(K)}\)}
\end{algorithm}
In this section, we present \sysname, our bridging framework for converting a pretrained autoregressive VLM into a same-architecture dVLM. \sysname\ contains two components: \textbf{Progressive Supervised Block Merging} and \textbf{Stage-wise dVLM Distillation}. The first component gradually enlarges the decoding block size during teacher-forcing post-training, so that the model does not need to jump directly from causal decoding to large-block diffusion. The second component uses a fixed small-block dVLM as the teacher to recover the performance drop caused by larger block sizes.

The overall idea is simple. Directly adapting a strong autoregressive VLM to a large-block dVLM usually causes a clear accuracy drop, because both the decoding pattern and the optimization difficulty change too abruptly. \sysname\ addresses this problem in two steps: it first builds a stable small-block diffusion anchor from the pretrained VLM, and then progressively enlarges the block size while repeatedly distilling from this anchor checkpoint.

In our main configuration, the stage-wise teacher-forcing training objective under block size \(B_k\) is the mixed-noise objective \(\mathcal L_{\mathrm{mix}}(\theta; B_k)\), which is defined later in Sec.~\ref{sec:mixed-noise}. Therefore, throughout the bridge, the corruption family and the same-position denoising target remain fixed, while only the block granularity changes across stages. Algorithm~\ref{alg:pbm} summarizes the full procedure.

\subsection{Progressive Supervised Block Merging}
\label{sec:pbm}

We use the block schedule
\begin{equation}
(B_1,B_2,B_3,B_4) = (4,8,16,32).
\label{eq:bridge-schedule}
\end{equation}
The first stage converts the pretrained autoregressive VLM into a small-block dVLM with block size \(B_1=4\). We denote the resulting checkpoint by \(\theta^{(1)}\), and use it as the \textbf{anchor model} throughout the rest of \sysname. This checkpoint is important because it already operates in the diffusion regime, while still remaining close to the original autoregressive model and preserving most of its capability.

For each later stage \(k \ge 2\), we enlarge the block size by merging adjacent blocks from the previous stage:
\begin{equation}
B_k = 2 B_{k-1}.
\label{eq:block-merge}
\end{equation}
In this sense, stage \(k\) does not introduce a completely new decoding pattern; it simply merges two neighboring blocks from the previous stage into one larger block. This makes the transition much smoother than directly training a large-block dVLM from scratch.

Across all stages, the mixed-noise corruption family and the same-position denoising target remain unchanged, and only the block granularity is modified. In particular, the model always reconstructs the clean token at the same position, rather than predicting the next token as in autoregressive training. This consistent position-aligned supervision makes it easier for the model to reuse the token prediction and multimodal reasoning patterns learned at smaller blocks.

\subsection{Stage-wise dVLM Distillation}
\label{sec:distillation}

Although progressive block merging makes the transition to larger block sizes much smoother, larger blocks still introduce a noticeable performance drop in practice. To reduce this drop, after each stage-\(k\) training phase we run a separate distillation phase from the fixed anchor teacher \(\theta^{(1)}\).

\begin{figure}[t]
    \centering
    \includegraphics[width=0.90\columnwidth]{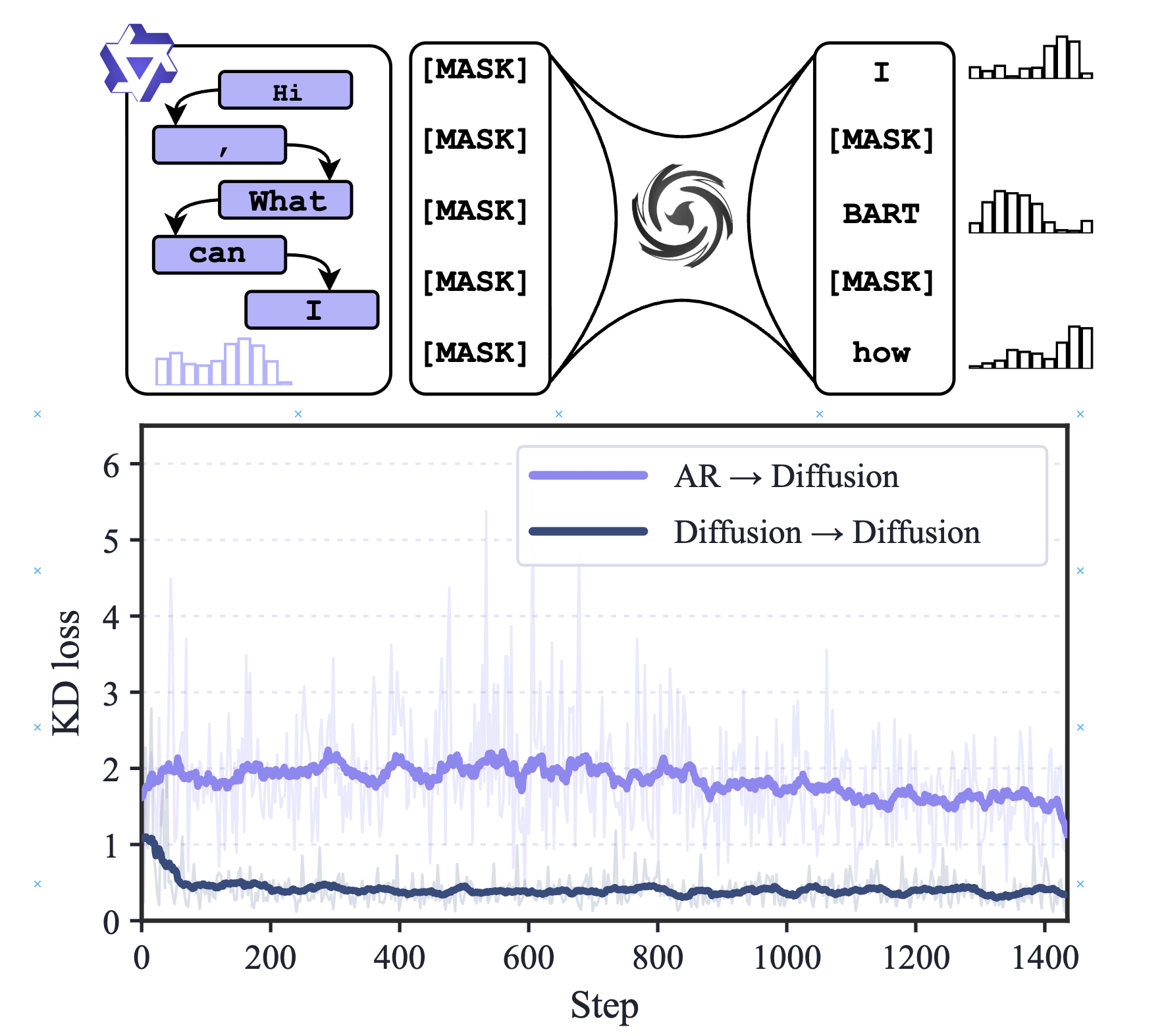}
    \caption{The autoregressive VLM produces next-token logits under clean causal prefix states, while the dVLM predicts same-position logits under corrupted diffusion states. This mismatch makes direct AR-to-dVLM KL supervision poorly aligned.}
    \label{fig:cross-stage_dvlm_distill}
\end{figure}

We intentionally keep the teacher fixed to the small-block anchor instead of using the original autoregressive VLM or the previous large-block checkpoint. As illustrated in Figure~\ref{fig:cross-stage_dvlm_distill}, direct autoregressive-to-diffusion distillation is poorly aligned: the autoregressive VLM predicts the next token under clean causal prefix states, whereas the diffusion student performs same-position denoising under corrupted diffusion states. The two logits are therefore not directly comparable, which leads to a larger KL mismatch in practice. In contrast, the anchor checkpoint \(\theta^{(1)}\) is already a strong diffusion model and provides a stable reference for all later stages.

Formally, let \(\mathcal S_t \subseteq \mathcal C\) denote the positions supervised by the mixed-noise objective. Concretely, \(\mathcal S_t\) contains both masked positions and visible corrupted positions. After the stage-\(k\) checkpoint \(\tilde{\theta}^{(k)}\) is obtained, we continue optimization from \(\tilde{\theta}^{(k)}\) using the fixed teacher \(\theta^{(1)}\). Given the same training example \((Q,x_1)\) and the same noised sequence \(x_t \sim q(x_t\mid x_1,t)\), the teacher and the student produce token logits \(z_{T,i}\) and \(z_{S,i}\) for each \(i\in\mathcal S_t\). We then minimize
\begin{equation}
\begin{aligned}
\mathcal L_{\mathrm{kd}}^{(k)} ={} & \mathbb E_{(Q,x_1),x_t,t} \Bigg[ \frac{1}{\max(1,|\mathcal S_t|)} \\
& \sum_{i\in\mathcal S_t} \tau^2\, \mathrm{KL} \Big( \mathrm{softmax}(z_{T,i}/\tau) \;\|\; \mathrm{softmax}(z_{S,i}/\tau) \Big) \Bigg],
\end{aligned}
\label{eq:stage-kd}
\end{equation}
where \(\tau\) is the distillation temperature.

This distillation phase does not inject new external knowledge. Its role is to preserve the diffusion capability of the anchor model while adapting the student to harder, larger-block decoding conditions. A similar motivation has been explored in sequential RL post-training, where stage-wise distillation is used to recover or preserve capabilities acquired in earlier stages \cite{zeng2026glm5}. In our setting, the challenge comes from increasing block size rather than changing RL objectives. Progressive supervised block merging handles the smooth transition of block granularity, and stage-wise dVLM distillation compensates for the performance loss introduced by larger blocks. Together, they form the core mechanism of \sysname.
\section{\modname}

\begin{figure*}[t]
    \centering
    \includegraphics[width=0.95\linewidth]{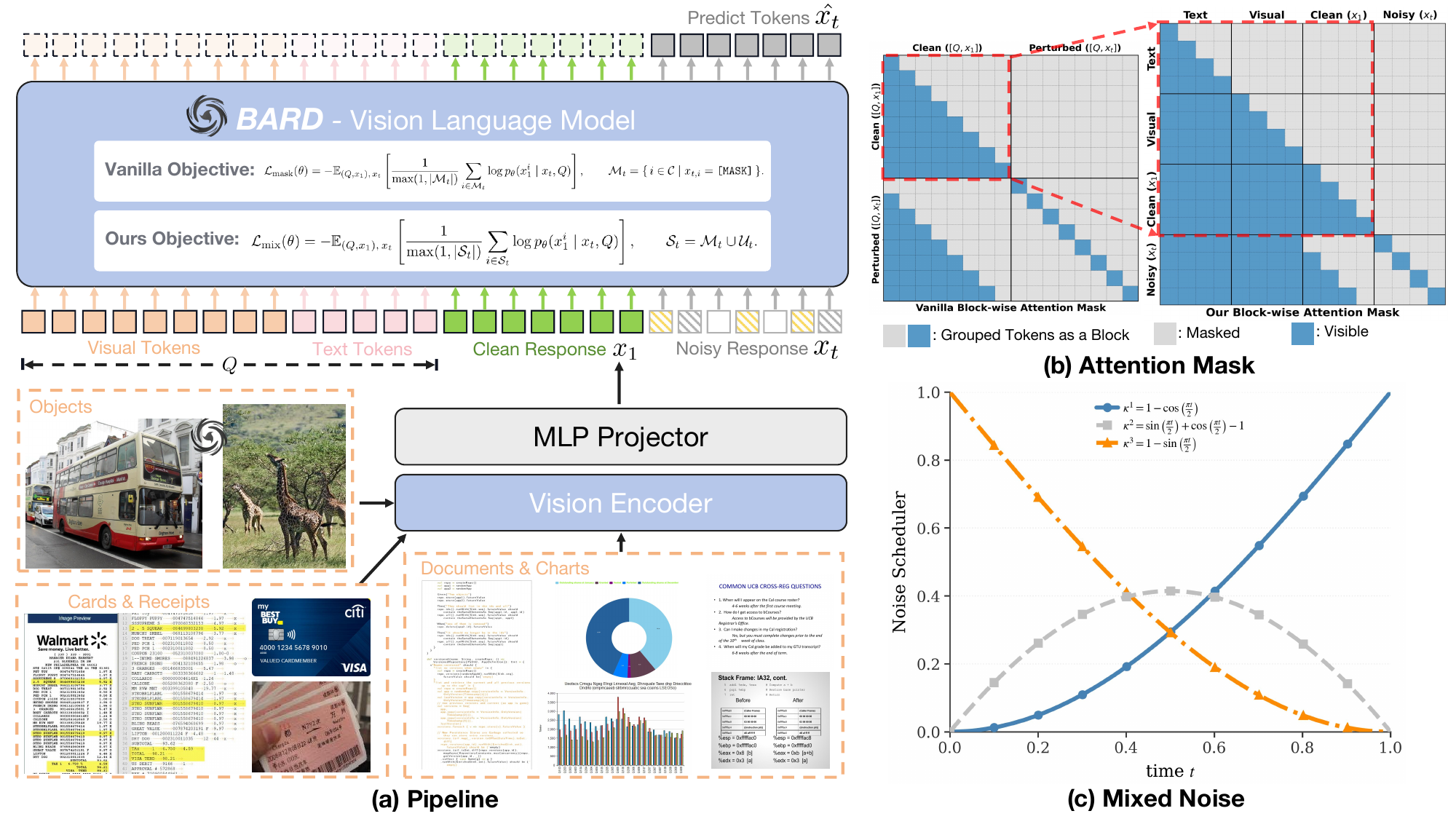}
    \caption{Overview of \modname.
    (a) Pipeline. We use the packed training layout \([Q, x_1, x_t]\), where the multimodal context \(Q\), the clean response \(x_1\), and the corrupted response \(x_t\) are processed in a single sequence. The two objective boxes correspond to the standard masked diffusion loss \(\mathcal L_{\mathrm{mask}}\) and the mixed-noise loss \(\mathcal L_{\mathrm{mix}}\).
    (b) Attention mask. Compared with the vanilla block-wise mask, our packed layout removes the redundant duplication of \(Q\). The \(i\)-th noisy block \(x_t^i\) attends only to the shared context \(Q\), the causal prefix \(\{x_1^0,\dots,x_1^{i-1}\}\), and tokens inside the same noisy block.
    (c) Mixed-noise scheduler. The coefficients \(\kappa_t^1,\kappa_t^2,\kappa_t^3\) control the interpolation among clean tokens, uniformly corrupted visible tokens, and masked tokens.}
    \label{fig:architecture}
\end{figure*}

In this section, we describe the practical training design used in the main \modname\ configuration. It includes two components: a mixed-noise scheduler and a memory-friendly packed training layout. The mixed-noise scheduler defines the stage-wise teacher-forcing objective used in Sec.~\ref{sec:method}, while the packed layout reduces the memory cost of multimodal diffusion training.

\subsection{Mixed-Noise Scheduler}
\label{sec:mixed-noise}

Standard absorbing-state diffusion corrupts completion tokens only through the special token \texttt{[MASK]}. Given multimodal context \(Q\) and clean completion \(x_1 \in \mathcal{V}^L\), let
\begin{equation}
\mathcal M_t = \{\, i \in \mathcal C \mid x_{t,i}=\texttt{[MASK]} \,\}
\label{eq:mask_set}
\end{equation}
be the set of masked positions. The standard masked diffusion objective can then be written as
\begin{equation}
\mathcal L_{\mathrm{mask}}(\theta)
=
-
\mathbb E_{(Q,x_1),\,x_t}
\left[
\frac{1}{\max(1,|\mathcal M_t|)}
\sum_{i\in\mathcal M_t}
\log p_\theta(x_1^i \mid x_t,Q)
\right].
\label{eq:l_mask}
\end{equation}
This objective teaches the model to recover missing tokens, but it does not train the model to correct visible but incorrect tokens.

To make the decoder revisable, we additionally inject random visible-token corruption into the completion. Let \(x_0\) denote the source sequence whose completion tokens are all set to \texttt{[MASK]}, and let \(x_1\) denote the clean completion. For each completion position \(i\), we sample the corrupted token from
\begin{equation}
q(x_t^i \mid x_0^i, x_1^i, t)
=
\kappa_t^1 \delta_{x_1^i}(x_t^i)
+
\kappa_t^2 p_u(x_t^i)
+
\kappa_t^3 \delta_{x_0^i}(x_t^i),
\label{eq:mix_corruption}
\end{equation}
where \(p_u\) is the uniform distribution over vocabulary tokens except \texttt{[MASK]}, and \(\delta_a(\cdot)\) is a point mass at token \(a\). We use
\begin{align}
\kappa_t^1 &= 1 - \cos(\pi t/2), \\
\kappa_t^2 &= \cos(\pi t/2) + \sin(\pi t/2) - 1, \\
\kappa_t^3 &= 1 - \sin(\pi t/2),
\end{align}
with \(t \sim \mathrm{U}(0,1)\), so that $\kappa_t^1 + \kappa_t^2 + \kappa_t^3 = 1$.
Here \(\kappa_t^3\) is the standard mask branch, while \(\kappa_t^2\) introduces visible-token corruption. Let \(\mathcal U_t\) be the set of uniformly corrupted visible positions, and define
\begin{equation}
\mathcal S_t = \mathcal M_t \cup \mathcal U_t.
\label{eq:mix_supervised_set}
\end{equation}
We then optimize
\begin{equation}
\mathcal L_{\mathrm{mix}}(\theta)
=
-
\mathbb E_{(Q,x_1),\,x_t}
\left[
\frac{1}{\max(1,|\mathcal S_t|)}
\sum_{i\in\mathcal S_t}
\log p_\theta(x_1^i \mid x_t,Q)
\right].
\label{eq:mix_loss}
\end{equation}
Compared with \(\mathcal L_{\mathrm{mask}}\), \(\mathcal L_{\mathrm{mix}}\) keeps the same same-position denoising target, but extends the supervised set from only masked positions to both masked and visibly corrupted positions.

At inference time, \modname\ uses a simple confidence-based refinement rule. Given the current state \(x^{(s)}\), the model predicts
\begin{equation}
\hat{x}_i^{(s)}=\arg\max_{v\in\mathcal V} p_\theta(x_i=v\mid x^{(s)},Q),
c_i^{(s)}=\max_{v\in\mathcal V} p_\theta(x_i=v\mid x^{(s)},Q).
\label{eq:mix_confidence}
\end{equation}
A masked token is materialized once its confidence exceeds a threshold \(\eta\). For an already visible token, we allow revision only when a later forward pass produces a higher-confidence prediction. In this way, high-confidence tokens stabilize early, while low-confidence tokens remain revisable.

\subsection{Memory-Friendly Training}
\label{sec:packed-training}

Multimodal diffusion training is memory-heavy because visual tokens usually dominate the sequence length. A naive implementation duplicates the multimodal context \(Q\) to form a clean branch \([Q,x_1]\) and a noisy branch \([Q,x_t]\). This doubles the visual context and sharply reduces the usable context window.

To avoid this redundancy, we use a packed sequence layout
\begin{equation}
\tilde{x} = [Q,\; x_1,\; x_t].
\label{eq:packed-sequence}
\end{equation}
Here the multimodal context \(Q\) is stored only once, followed by the clean response \(x_1\) and the corrupted response \(x_t\). This layout allows all diffusion blocks to be trained in a single forward pass.

The packed layout requires a matching attention mask. Tokens in \(Q\) and \(x_1\) keep the original causal mask of the pretrained backbone. For the \(i\)-th noisy block \(x_t^i\), the allowed attention set is
\begin{equation}
\mathrm{Attn}(x_t^i)
=
Q \;\cup\; \{x_1^0,\dots,x_1^{i-1}\} \;\cup\; x_t^i.
\label{eq:packed-attn}
\end{equation}
That is, the current noisy block can attend to the shared context, the causal prefix of the clean response, and noisy tokens inside the same block, but not to future clean blocks or future noisy blocks. This prevents information leakage while preserving blockwise parallel training.
This design is mainly an efficiency improvement. It does not change the diffusion objective itself, but it makes multimodal diffusion training much more practical by removing redundant copies of \(Q\), reducing memory cost, and preserving a larger effective context window.

\section{Experiment}

\newcommand{\exptablesetup}[1]{%
    \footnotesize
    \setlength{\tabcolsep}{#1}%
    \renewcommand{\arraystretch}{1.08}%
}

\begin{table*}[t]
\centering
\caption{Comparison on seven multimodal benchmarks spanning reasoning, general VQA, and document understanding. For \modname, we report the final large-block checkpoint evaluated at \(B=4~\text{and}~32\).}
\label{tab:vlm_performance}
\setlength{\tabcolsep}{3pt}
\begin{tabular}{@{}lcccccccc@{}}
\toprule
Model & Parameters  & MMMU$_{\text{val}}$ & MMMU-Pro$_{\text{standard}}$ & MME$_{\text{sum}}$ & RealWorldQA & MMStar & AI2D & ChartQA \\
\midrule
\multicolumn{9}{c}{\textbf{AutoRegressive Vision-Language Models}} \\
\midrule
Qwen3-VL~\cite{bai2025qwen3vl}    & 4B  & 47.9 & 35.0 & 2297 & 70.5 & 56.9 & 81.0 & 80.9 \\
Qwen3-VL~\cite{bai2025qwen3vl}    & 8B  & 53.0 & 36.0 & 2379 & 69.5 & 59.9 & 83.5 & 84.0 \\
InternVL3.5~\cite{wang2025internvl3} & 4B  & 57.4 & 38.2 & 2236 & 66.7 & 65.6 & 80.6 & 86.2 \\
InternVL3.5~\cite{wang2025internvl3} & 8B  & 57.2 & 41.0 & 2359 & 63.1 & 66.3 & 82.1 & 87.0 \\
\midrule
\multicolumn{9}{c}{\textbf{Diffusion Vision-Language Models}} \\
\midrule
LLaDA-V~\cite{you2025llada}     & 8B  & 48.8 & 35.4 & 1998 & 63.4 & 60.4 & 77.8 & 78.2 \\
Dream-VL~\cite{ye2025dream}    & 7B  & 51.6 & 25.0 & 2179 & 67.7 & 59.9 & 80.4 & 86.2 \\
LaviDa~\cite{li2025lavida}      & 8B  & 44.2 & 28.6 & 1711 & 40.3 & 47.0 & 70.1 & 64.6 \\
SDAR-VL~\cite{cheng2025sdar}     & 8B  & 44.0 & 28.2 & 2142 & 66.1 & 53.3 & 79.6 & 82.4 \\
MMaDA~\cite{yang2025mmada}       & 8B  & 30.2 & 21.5 & 1287 & 28.2 & 25.7 & 54.9 & 43.2 \\
Dimple-VL~\cite{yu2505dimple}   & 7B  & 46.4 & 24.1 & 1924 & 51.9 & 47.7 & 74.2 & 58.4 \\
\midrule
\multicolumn{9}{c}{\textbf{\modname\ Converted from InternVL3.5}} \\
\midrule
InternVL3.5~\cite{wang2025internvl3} & 4B  & 57.4 & 38.2 & 2236 & 66.7 & 65.6 & 80.6 & 86.2 \\
BARD-InternVL3.5 (B=4) & 4B & \textbf{57.4} & \textbf{38.6} & \textbf{2287} & 66.2 & 65.1 & \textbf{80.9} & 85.6 \\
\midrule
\multicolumn{9}{c}{\textbf{\modname\ Converted from Qwen3-VL}} \\
\midrule
Qwen3-VL~\cite{bai2025qwen3vl} & 4B & 47.9 & \textbf{35.0} & 2297 & 70.5 & 56.9 & 81.0 & 80.9 \\
Qwen3-VL-SFT & 4B & 49.5 & 33.9 & 2300 & 71.9 & 59.5 & 81.7 & \textbf{81.0} \\
\textbf{BARD-Qwen3-VL $(B=32)$ } & 2B & 42.0 & 27.9 & 2045 & 64.6 & 53.1 & 72.6 & 76.8 \\
\textbf{BARD-Qwen3-VL $(B=32)$ } & 4B & 53.0 & 34.2 & 2305 & 71.9 & 63.6 & 82.8 & 80.2 \\
\textbf{BARD-Qwen3-VL $(B=4)$ } & 8B & 54.6 & 37.6 & 2393 & 70.7 & 65.0 & 83.2 & 84.6 \\
\bottomrule
\end{tabular}
\end{table*}

\subsection{Setups}
\paragraph{\textbf{Training}}
All experiments are conducted using publicly available datasets. We use LLaVA-OneVision-1.5~\cite{an2025llava} and FineVision~\cite{wiedmann2025finevision} for supervised adaptation. LLaVA-OneVision-1.5 provides diverse multimodal instruction-following data for general vision-language reasoning, while FineVision contributes higher-quality visual reasoning and comprehension annotations. 
Because both sources are aggregated from many underlying subsets, with substantial overlap between them, we first deduplicate the shared subsets during data preprocessing. We then filter out samples whose shorter image side is smaller than the visual patch size (\(16\)), whose aspect ratio exceeds \(100\), or whose total sequence length is longer than \(16\)K tokens.
For stage-wise distillation, we randomly sample 300 examples from each subset to construct a distilled dataset of approximately 100K instances.

All models are trained on 8 NVIDIA H200 GPUs with a global batch size of 256. We jointly optimize all parameters using AdamW with beta coefficients $(0.9, 0.95)$. The language model uses a peak learning rate of $2\times 10^{-5}$, while the vision tower uses $2\times 10^{-6}$. We apply a linear warmup for the first 1,000 steps followed by cosine decay to $1\times 10^{-6}$, and use gradient clipping with a maximum norm of 1.0 for stability. For stage-wise dVLM distillation, we use a distillation temperature of \(\tau=1.0\) in all reported experiments.

\paragraph{\textbf{Evaluation}}
We evaluate our models on seven benchmarks spanning three representative categories of multimodal understanding: multimodal reasoning (MMMU$_{\mathrm{val}}$, MMMU-Pro$_{\mathrm{standard}}$, and MME), general visual question answering (RealWorldQA and MMStar), and document understanding (AI2D and ChartQA). All benchmark evaluations are conducted with VLMEvalKit, the open-source evaluation toolkit from OpenCompass, and we align the evaluation settings with those reported in Qwen and other baseline papers to ensure fair comparison, and run all metric evaluations on 4 NVIDIA RTX 4090 GPUs. In the main comparison, we include both state-of-the-art autoregressive VLMs and recent diffusion VLMs, and evaluate \modname\ at 2B, 4B and 8B scales.


\subsection{Comparison with State-of-the-Art VLMs}
\begin{figure}[ht]
    \centering
    \includegraphics[width=0.95\linewidth]{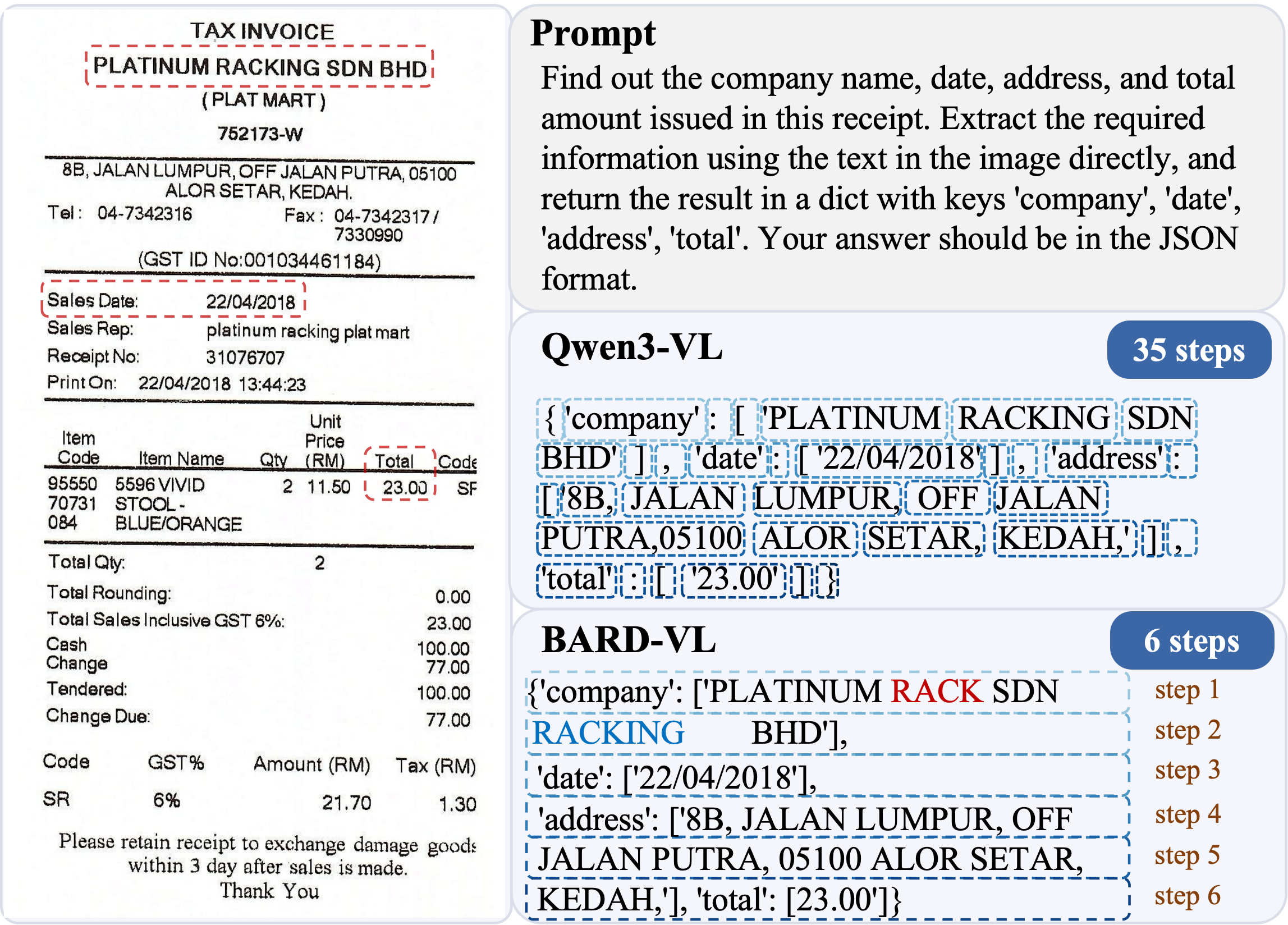}
    \caption{\textbf{Comparison on document understanding.}}
    \label{fig:ocrbench-case}
\end{figure}

Table~\ref{tab:vlm_performance} reports the main comparison against strong autoregressive and diffusion vision-language models. When converted from Qwen3-VL, \modname\ preserves the strengths of the autoregressive backbone and improves it on most benchmarks. At the 4B scale, \modname\ improves over Qwen3-VL-4B on five of the seven benchmarks, with gains of +5.1 on MMMU$_{\mathrm{val}}$, +8 on MME$_{\mathrm{sum}}$, +1.4 on RealWorldQA, +6.7 on MMStar, and +1.8 on AI2D, while remaining close on MMMU-Pro$_{\mathrm{standard}}$ (-0.8) and ChartQA (-0.7). At the 8B scale, \modname\ improves over Qwen3-VL-8B on six of seven benchmarks, including +1.6 on MMMU$_{\mathrm{val}}$, +1.6 on MMMU-Pro$_{\mathrm{standard}}$, +14 on MME$_{\mathrm{sum}}$, +1.2 on RealWorldQA, +5.1 on MMStar, and +0.6 on ChartQA, with only a small drop on AI2D (-0.3). These results show that a strong autoregressive VLM can be adapted to diffusion decoding while largely preserving, and often improving, multimodal capability. Figure~\ref{fig:ocrbench-case} further illustrates this advantage on a receipt information extraction task. While \modname\ remains broadly comparable to Qwen3-VL, it reaches the structured prediction in only 6 diffusion refinement steps, compared with 35 autoregressive decoding steps for Qwen3-VL, indicating substantially better decoding parallelism for long structured outputs.

Compared with prior diffusion VLMs, \modname\ is consistently competitive and usually stronger across the reported suite. \modname\ 8B outperforms LLaDA-V-8B on all seven benchmarks, and \modname\ 4B also surpasses Dimple-VL on all seven. Relative to Dream-VL and SDAR-VL, \modname\ 4B wins on six of seven benchmarks, with the only exception being ChartQA. Overall, these comparisons show that the proposed conversion recipe substantially narrows the quality gap between diffusion VLMs and strong autoregressive baselines.

\subsection{Ablations}

\begin{table*}[ht]
\centering
\caption{Ablation study on block sizes and initialization strategies.}
\label{tab:block_size_scaling}
\begin{tabular}{@{}clccccccc@{}}
\toprule
\textbf{Block Size} & \textbf{Strategy} & MMMU$_{\mathrm{val}}$ & MMMU-Pro$_{\mathrm{standard}}$ & MME$_{\mathrm{sum}}$ & RealWorldQA & MMStar & AI2D & ChartQA \\
\midrule
4  & Direct            & 52.8 & 34.9 & 2359 & 72.3 & 63.6 & 82.3 & 83.7 \\
\midrule
\multirow{2}{*}{8}  & Direct             & 50.0 & 33.2 & 2331 & 70.2 & 61.0 & 81.2 & 82.0 \\
                    & Warm-start (BS=4)  & 52.2 {\scriptsize\textcolor{DarkGreen}{\textbf{+2.2}}} & 34.7 {\scriptsize\textcolor{DarkGreen}{\textbf{+1.5}}} & 2371 {\scriptsize\textcolor{DarkGreen}{\textbf{+40}}} & 72.0 {\scriptsize\textcolor{DarkGreen}{\textbf{+1.8}}} & 64.3 {\scriptsize\textcolor{DarkGreen}{\textbf{+3.3}}} & 82.3 {\scriptsize\textcolor{DarkGreen}{\textbf{+1.1}}} & 82.6 {\scriptsize\textcolor{DarkGreen}{\textbf{+0.6}}} \\
\midrule
\multirow{2}{*}{16} & Direct             & 47.8 & 32.5 & 2324 & 69.5 & 58.7 & 80.1 & 82.0 \\
                    & Warm-start (BS=8)  & 51.5 {\scriptsize\textcolor{DarkGreen}{\textbf{+3.7}}} & 34.2 {\scriptsize\textcolor{DarkGreen}{\textbf{+1.7}}} & 2347 {\scriptsize\textcolor{DarkGreen}{\textbf{+23}}} & 70.7 {\scriptsize\textcolor{DarkGreen}{\textbf{+1.2}}} & 60.9 {\scriptsize\textcolor{DarkGreen}{\textbf{+2.2}}} & 82.0 {\scriptsize\textcolor{DarkGreen}{\textbf{+1.9}}} & 82.5 {\scriptsize\textcolor{DarkGreen}{\textbf{+0.5}}} \\
\midrule
\multirow{2}{*}{32} & Direct             & 49.8 & 32.3 & 2195 & 67.2 & 54.9 & 78.4 & 77.3 \\
                    & Warm-start (BS=16) & 50.0 {\scriptsize\textcolor{DarkGreen}{\textbf{+0.2}}} & 32.9 {\scriptsize\textcolor{DarkGreen}{\textbf{+0.6}}} & 2262 {\scriptsize\textcolor{DarkGreen}{\textbf{+67}}} & 68.2 {\scriptsize\textcolor{DarkGreen}{\textbf{+1.0}}} & 55.9 {\scriptsize\textcolor{DarkGreen}{\textbf{+1.0}}} & 79.8 {\scriptsize\textcolor{DarkGreen}{\textbf{+1.4}}} & 80.4 {\scriptsize\textcolor{DarkGreen}{\textbf{+3.1}}} \\
\bottomrule
\end{tabular}
\end{table*}

\begin{table*}[ht]
\centering
\caption{Performance comparison across different training dataset sizes for the \modname\ 4B anchor model at block size \(B=4\).}
\label{tab:dataset_comparison}
\setlength{\tabcolsep}{3pt}
\begin{tabular}{@{}cccccccc@{}}
\toprule
\textbf{Dataset Size} & MMMU$_{\mathrm{val}}$ & MMMU-Pro$_{\mathrm{standard}}$ & MME$_{\mathrm{sum}}$ & RealWorldQA & MMStar & AI2D & ChartQA \\
\midrule
110K & 49.5 & 33.4 & 2225 & 68.7 & 56.4 & 80.8 & 82.0 \\
220K & 51.3 & 34.7 & 2288 & 70.8 & 60.2 & 81.6 & 82.3 \\
440K & 51.1 & \textbf{35.3} & 2275 & 71.2 & 60.6 & \textbf{82.5} & 83.0 \\
2.2M & 51.6 & 34.3 & 2306 & 71.8 & 61.3 & 81.9 & 83.6 \\
4.4M & \textbf{52.8} & 34.9 & \textbf{2359} & \textbf{72.3} & \textbf{63.6} & 82.3 & \textbf{83.7} \\
\bottomrule
\end{tabular}
\end{table*}
\paragraph{\textbf{Dataset Size}}
We first study how the scale of supervised adaptation data affects \modname. To isolate the effect of data scale, we vary the number of training examples from 110K to 4.4M while keeping the model architecture and optimization setting fixed. This ablation asks two questions: whether the proposed bridge remains effective in the low-data regime, and whether its gains continue to improve with larger-scale multimodal supervision. 
Table~\ref{tab:dataset_comparison} reveals two main findings. First, \modname\ is already effective with limited supervision. Even with only 110K examples, the model reaches 49.5 on MMMU, 68.7 on RealWorldQA, and 82.0 on ChartQA, showing that the proposed bridge does not require massive adaptation data to become competitive. Second, increasing the data scale leads to steady overall improvement, with the 4.4M setting achieving the best results on five of the seven reported benchmarks, including MMMU, MME, RealWorldQA, MMStar, and ChartQA. The largest gains appear on MME and MMStar, indicating that these capabilities benefit most clearly from additional supervision.

At the same time, the scaling trend is not perfectly monotonic on every benchmark. MMMU-Pro and AI2D peak at 440K and remain roughly saturated afterward, while the remaining benchmarks continue to improve at larger scales. 
It suggests that the proposed diffusion adaptation recipe is both data-efficient and scalable.
A modest amount of supervised data is sufficient to obtain a strong dVLM, while larger-scale supervision further improves broad multimodal capability across reasoning, general VQA, and document understanding benchmarks.

\begin{table*}[htbp]
\centering
\caption{Ablation of cross-stage dVLM distillation on \modname 4B at target block sizes \(B \in \{8,16,32\}\). The diffusion teacher is the fixed small-block anchor checkpoint \(\theta^{(1)}\).}
\label{tab:ablation_block_policy}
\setlength{\tabcolsep}{3pt}
\begin{tabular}{@{}lclccccccc@{}}
\toprule
\textbf{Model} & \textbf{Block Size} & \textbf{Strategy} & MMMU$_{\mathrm{val}}$ & MMMU-Pro$_{\mathrm{standard}}$ & MME$_{\mathrm{sum}}$ & RealWorldQA & MMStar & AI2D & ChartQA \\
\midrule
\multirow{3}{*}{\modname\ 4B} & \multirow{3}{*}{8}
& w/o distillation  & 52.2 & 34.7 & 2371 & 72.0 & 64.3 & 82.3 & \textbf{82.6} \\
& & AR distillation & 51.2 & 34.9 & \textbf{2379} & 71.1 & 62.9 & 82.2 &  81.3 \\
& & Diffusion distillation & \textbf{53.9} & \textbf{37.5} & 2376 & \textbf{72.8} & \textbf{64.9} & \textbf{83.1} & 81.1 \\
\midrule
\multirow{3}{*}{\modname\ 4B} & \multirow{3}{*}{16}
& w/o distillation     & 51.5 & 34.2 & 2347 & 70.7 & 60.9 & 82.0 & \textbf{82.5} \\
& & AR distillation    & 51.6 & 34.0 & \textbf{2349} & 70.3 & 54.5 & 81.6 & 81.0 \\
& & Diffusion distillation & \textbf{53.6} & \textbf{36.1} & 2340 & \textbf{72.5} & \textbf{64.1} & \textbf{83.0} & 81.0 \\
\midrule
\multirow{3}{*}{\modname\ 4B} & \multirow{3}{*}{32}
& w/o distillation     & 50.0 & 32.9 & 2262 & 68.2 & 55.9 & 79.8 & 80.4 \\
& & AR distillation    &  49.7 & 31.8 & 2252 & 68.6 & 52.1 & 80.0 & 81.0 \\
& & Diffusion distillation & \textbf{53.0} & \textbf{34.2} & \textbf{2305} & \textbf{71.9} & \textbf{63.6} & \textbf{82.8} & \textbf{80.2} \\
\bottomrule
\end{tabular}
\end{table*}

\begin{table*}[htbp]
\centering
\caption{Ablation of noise type for \modname\ 4B with block size \(B=4\).}
\label{tab:noise_scheduler}
\begin{tabular}{@{}l c c c c c c c@{}}
\toprule
Noise Type & MMMU$_{\mathrm{val}}$ & MMMU-Pro$_{\mathrm{standard}}$ & MME$_{\mathrm{sum}}$ & RealWorldQA & MMStar & AI2D & ChartQA \\
\midrule
Mask    & 52.8 & 34.9 & 2359 & 72.3 & 63.6 & 82.3 & 83.7 \\
Uniform & 50.5 & 32.3 & \textbf{2384} & 71.6 & \textbf{63.8} & 80.8 & 84.0 \\
Mixture & \textbf{52.8} & \textbf{35.1} & 2366 & \textbf{73.4} & 63.3 & \textbf{82.3} & \textbf{84.3} \\
\bottomrule
\end{tabular}
\end{table*}

\paragraph{\textbf{Progressive Block Merging}}
To understand the role of progressive block merging, we vary the target decoding block size and compare two training strategies: direct adaptation at the target block size, and stage-wise warm-starting from the checkpoint trained at the immediately preceding smaller block. Table~\ref{tab:block_size_scaling} reports the results.

Direct adaptation becomes increasingly difficult as the target block size grows. Relative to the \(B=4\) model, direct training at \(B=8\), \(16\), and \(32\) degrades performance across the full evaluation suite, indicating that large-block diffusion adaptation introduces a substantial capability-preservation challenge.

Stage-wise warm-starting consistently narrows this gap. At every target block size, the warm-started model outperforms direct adaptation on all seven benchmarks. At \(B=8\), warm-starting improves MMMU from 50.0 to 52.2, MME from 2331 to 2371, and MMStar from 61.0 to 64.3, while even slightly surpassing the \(B=4\) model on MME and MMStar. The \(B=8\) results show that the larger block size is manageable when introduced progressively, whereas direct adaptation causes most of the degradation.
Pushing the block size further to \(B=16\) and \(B=32\) makes the task harder, and some performance loss remains even with warm-starting. Even so, progressive scaling still recovers a substantial portion of the gap over direct adaptation, including +3.7 on MMMU and +2.2 on MMStar at \(B=16\), and +67 on MME and +3.1 on ChartQA at \(B=32\).


We further evaluate the stage-wise dVLM distillation introduced in Sec.~\ref{sec:distillation} to test whether a small-block anchor dVLM teacher can recover the capability lost when the student is pushed to a larger block size. For each target student block size \(B \in \{8,16,32\}\), we compare three training strategies: no distillation, distillation from the corresponding autoregressive VLM, and distillation from the fixed small-block anchor dVLM \(\theta^{(1)}\). All variants use the same data, model size, and optimization setting, and the KL loss is applied on the supervised positions defined by the chosen diffusion training objective under matched multimodal inputs. In our main setting, this includes both masked positions and uniformly corrupted visible positions. Table~\ref{tab:ablation_block_policy} reports the results.

The results show that stage-wise dVLM distillation is consistently more effective than autoregressive distillation, and that its value becomes larger as the target block size increases. At \(B=8\), diffusion distillation improves 6 of 7 benchmarks over training without distillation, including +1.7 on MMMU, +2.8 on MMMU-Pro, and +0.8 on RealWorldQA, whereas autoregressive distillation helps on only 2 of 7 metrics and reduces MMStar by 1.4 points. At \(B=16\), diffusion distillation again improves 5 of 7 benchmarks, with gains of +2.1 on MMMU, +1.8 on RealWorldQA, and +3.2 on MMStar, while autoregressive distillation brings little overall benefit and sharply degrades MMStar (-6.4). The advantage is most pronounced at \(B=32\), where diffusion distillation recovers a substantial portion of the large-block degradation, improving MMMU by +3.0, RealWorldQA by +3.7, MMStar by +7.7, and AI2D by +3.0 over the no-distillation baseline. Although the gains on ChartQA are limited, the overall trend is clear: stage-wise dVLM distillation provides a more compatible teacher signal for large-block diffusion students than autoregressive distillation, which supports our claim that matched diffusion supervision is important for preserving capability at higher block sizes.

\paragraph{\textbf{Noise Scheduler}}
Finally, we ablate the noise scheduler to evaluate the design of mixed noise. Since mixed-noise training is used in the main \modname\ configuration, this ablation isolates its contribution by comparing three corruption schemes under the same \(B=4\) setting: pure mask noise, pure uniform noise, and the proposed mixture noise. The goal is to test whether introducing visible-token corruption can improve revision behavior without sacrificing the optimization stability of absorbing-mask diffusion. Table~\ref{tab:noise_scheduler} reports the results.

The proposed mixture noise achieves the best overall balance. It matches or exceeds the strongest baseline on five of the seven benchmarks, including the best results on MMMU-Pro, RealWorldQA, and ChartQA, while avoiding the substantial reasoning degradation of pure uniform noise. In other words, mixed noise provides a better compromise between the two extremes: it introduces correction-oriented supervision from uniform corruption, while retaining the reasoning performance of mask-based diffusion.

\section{Conclusion}

This paper shows that a strong autoregressive VLM can be converted into a diffusion VLM with substantially improved decoding parallelism without sacrificing multimodal capability. We present \sysname, a simple bridging framework that combines progressive block merging, stage-wise dVLM distillation, and mixed-noise diffusion scheduler. The key idea is to avoid an abrupt jump from causal decoding to large-block parallel denoising: we first build a stable small-block diffusion anchor from a pretrained VLM, then progressively enlarge the decoding block size and recover quality with matched diffusion distillation. This design turns autoregressive-to-diffusion adaptation from an unstable one-step conversion into a stable, scalable post-training recipe.

Experiments validate both the effectiveness of the resulting model family and the principles behind the bridge. Across seven multimodal benchmarks, \modname\ consistently outperforms prior open diffusion VLMs, and when initialized from Qwen3-VL it improves over the source autoregressive model on five of seven benchmarks at 4B and six of seven benchmarks at 8B. The ablations further show that the bridge is data-efficient, that progressive block merging is necessary for stable scaling to larger block sizes, and that mixed-noise scheduler provides a favorable balance between revision ability and reasoning performance. Taken together, these results suggest that revision-aware diffusion adaptation is a practical route to narrowing the quality gap between diffusion and autoregressive VLMs while moving toward more parallel multimodal generation.

\section{Limitations}

The current study provides a strong first validation of autoregressive-to-diffusion bridging for vision-language models, but there is still room for broader exploration. In particular, further experiments across additional domains would help better understand the full scope of the proposed framework. In addition, our current results are centered on Qwen3-VL, and it remains valuable to verify the same bridging recipe on a wider set of backbone families and scales.

{\small
\bibliographystyle{plain}
\bibliography{egbib}
}

\end{document}